\documentclass[runningheads]{llncs}
\usepackage{graphicx}
\usepackage{comment}
\usepackage{amsmath,amssymb} 
\usepackage{color}
\usepackage{algorithm}
\usepackage{algorithmicx}
\usepackage{algpseudocode}

\usepackage{multirow}
\usepackage{enumitem}

\begin{document}
\pagestyle{headings}
\mainmatter
\def\ECCVSubNumber{1168}  

\title{BS-NAS: Broadening-and-Shrinking One-Shot NAS with Searchable Numbers of Channels} 

\titlerunning{Abbreviated paper title}
%
\author{Zan Shen \and
Jiang Qian \and
Bojin Zhuang \and
Shaojun Wang \and
Jing Xiao}
\authorrunning{F. Author et al.}
%
\institute{
Ping An Technology, Shanghai, China\\
\email{\{SHENZAN416,QIANJIANG456,ZHUANGBOJIN232,WANGSHAOJUN851,XIAOJING661\}@pingan.com.cn}}
\maketitle

\begin{abstract}
One-Shot methods have evolved into one of the most popular methods in Neural Architecture Search (NAS) due to weight sharing and single training of a supernet. However, existing methods generally suffer from two issues: predetermined number of channels in each layer which is suboptimal; and model averaging effects and poor ranking correlation caused by weight coupling and continuously expanding search space.
To explicitly address these issues, in this paper, a Broadening-and-Shrinking One-Shot NAS (BS-NAS) framework is proposed, in which `broadening' refers to broadening the search space with a {\em spring block} enabling search for numbers of channels during training of the supernet; while `shrinking' refers to a novel {\em shrinking} strategy gradually turning off those underperforming operations. The above innovations broaden the search space for wider representation and then shrink it by gradually removing underperforming operations, followed by an evolutionary algorithm to efficiently search for the optimal architecture. Extensive experiments on ImageNet illustrate the effectiveness of the proposed BS-NAS as well as the state-of-the-art performance.

\end{abstract}

\section{Introduction}
Deep learning methods have achieved great success in different machine learning tasks. However, most classical architectures and efficient components require domain experts to spend a lot of time and energy on continuous trials and errors. Neural Architecture Search (NAS), which formulates architecture modifications as a search problem, has become an active topic recently. Early works~\cite{zoph2016neural,baker2016designing,pham2018efficient,zoph2018learning,real2017large,xie2017genetic,real2019regularized} achieve successes by using reinforcement learning and evolutionary algorithm, but they do suffer from substantial resource consumption and time costs.  Consequently, many of the subsequent works focus on methods that reduce this computational burden.

Some recent approachs~\cite{pham2018efficient,brock2018smash,bender2018understanding,liu2018darts,cai2018proxylessnas} focus on weight sharing strategies, with One-Shot being one of them, in which an over-parameterized supernet comprising all candidate architectures from a predefined search space is trained, and in the consequent search process, architectures inheriting weights from this supernet are evaluated and selected without training from scratch.

Despite significantly speeding up the search process, One-Shot NAS methods have always been plagued by its stability and weight coupling.
We delve into the One-Shot methods and speculate that the ranking correlation between architectures inheriting weights from the supernet and fully trained stand-alone ones is less satisfactory. The boundaries between superior and inferior architectures are unavoidably blurred, and performance is averaged due to non-discriminatory weight sharing. In addition, Li et al.~\cite{li2019improving} take a Bayesian perspective to infer that the gap between true parameter posterior and proxy posterior increases with the number of models contained in the supernet. While empirically, the wider the search space in the initial period is, the more chances there are to search for better models.

\begin{figure*}[t]
\centering
\includegraphics[width=4.8in]{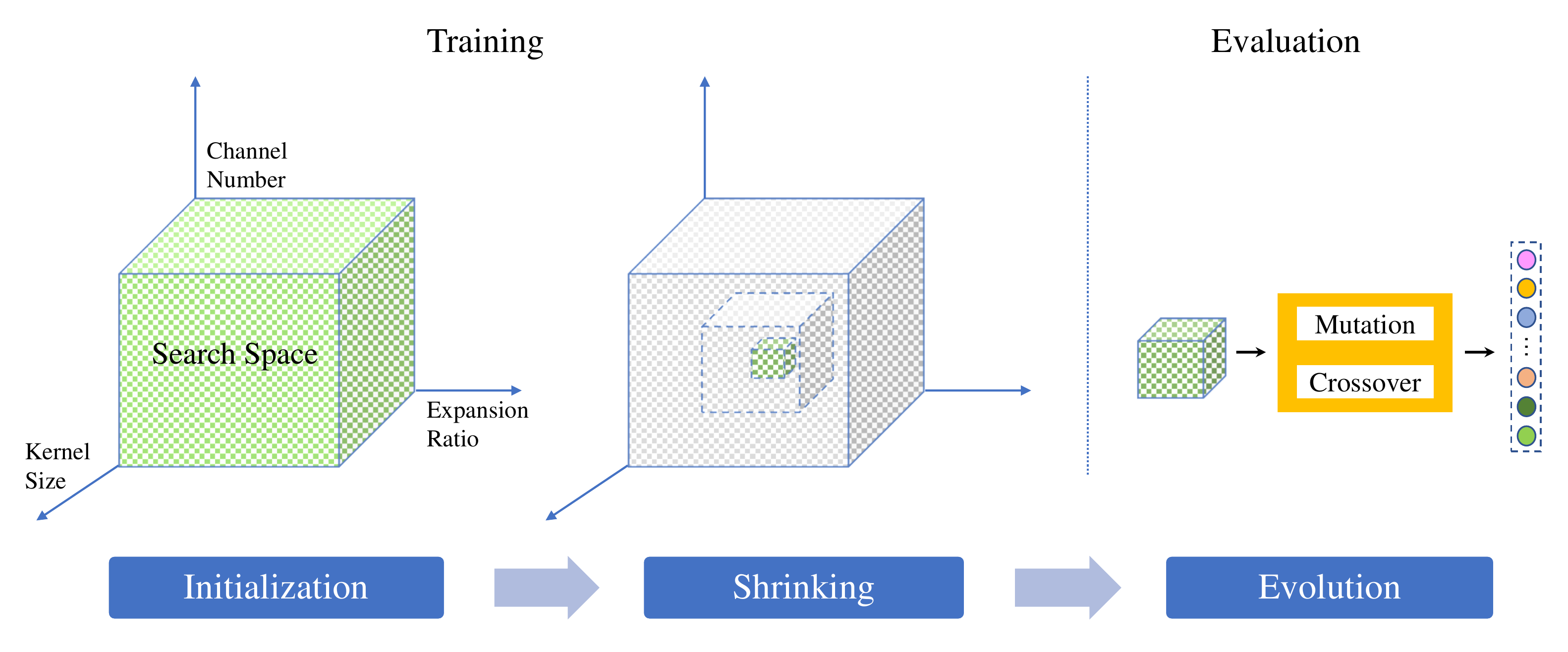}
\caption{An overview of the proposed BS-NAS framework. The green dots in the left figure represent optional operations in the `broadened' search space incorporating searchable channel numbers, the gray dots in the middle figure represent operations being eliminated, and the small cube in the right figure represents the `shrunk' search space on which an evolutionary algorithm is applied to search for the optimal neural architecture.}
\vspace{-1em}
\label{fig:fw} 
\end{figure*}

On the other side, the design of the search space forms a key component of neural architecture search. Previous methods~\cite{cai2018proxylessnas,chu2019scarletnas,mei2019atomnas} have incorporated the convolution type, kernel size, expansion ratio, network depth and other efficient modules (e.g. Squeeze-and-Excitation~\cite{hu2018squeeze}) into the search space of the supernet. As pointed out in the recent work of EfficientNet~\cite{tan2019efficientnet}, carefully balancing network depth, width, and resolution can lead to better performance. However, to the best of our knowledge, most existing One-Shot methods use manually predetermined numbers of channels during training.  Since the number of output channels in the current layer is correlated with the number of input channels in the next layer, it remains an intractable problem in One-Shot framework.

To address these issues, we propose a novel Broadening-and-Shrinking One-Shot NAS with searchable numbers of channels (BS-NAS), as illustrated in Figure.~\ref{fig:fw}. We break the conventional form that the number of channels in One-Shot method is artificially predefined and propose a supernet with searchable channel numbers. We further study the Inverted Residual Block of MobileNetV2~\cite{sandler2018mobilenetv2} and infer that, to achieve the optional number of channels, the numbers of output channels of the candidate blocks in the same layer must be an explicit value.
While such number of output channels depends only on the last linear pointwise convolution, and has nothing to do with the depthwise convolution which assumes the most important feature extraction. Based on this observation, we divide the network into several clusters and introduce a {\em spring block} which can flexibly adapt to different numbers of channels without disrupting the stability of the network. Then a novel {\em step-shrinking} strategy is proposed to effectively mitigate model averaging effects and facilitate search for optimal architecture. After a period of training, operations densely scattering on the supergraph are layer-wisely sorted and eliminated step by step. Only those operations that consistently perform well can survive the gradual elimination.
Finally, an evolutionary algorithm is applied to efficiently search for the optimal architecture on the shrunk supernet. It is worth mentioning that the above process is not limited to the MobileNetV2 framework and can be easily extended to other neural structures.

Our contributions can be summarized as follows:

\begin{enumerate}
  \item We propose a crafty {\em spring block} to broaden the search dimension of One-Shot methods, which enables search for number of channels during supernet training, and further enriches the representation of the search space.
  \item We propose a novel {\em step-shrinking} strategy to gradually eliminate those underperforming operations so as to mitigate model averaging effects and facilitate search for optimal architecture.
  \item Combining the above two strategies with an evolutionary algorithm, we propose a Broadening-and-Shrinking One-Shot NAS (BS-NAS) framework. Extensive experiments on ImageNet dataset illustrate the effectiveness of the proposed BS-NAS as well as the state-of-the-art performance.
\end{enumerate}

\section{Related Work}
\subsection{Neural Architecture Search}
Neural Architecture Search(NAS)  aims to generate a robust and well-performing
neural architecture by automatically selecting and combining different basic
operations from a predefined search space according to some
search strategies. Early works focused on reinforcement learning (RL)~\cite{zoph2016neural,baker2016designing,pham2018efficient,zoph2018learning} and  evolutionary algorithm (EA)~\cite{stanley2002evolving,real2017large,xie2017genetic,real2019regularized}. These works search on proxy task, candidate architectures are trained stand-alone, the performance  of which on the proxy task
 is used to evaluate the actual performance. However, such methods are both time-consuming and computationally expensive, especially on large-scale tasks (e.g., ImageNet).

\subsection{One-Shot NAS}
 To save resource and time, some recent works~\cite{pham2018efficient,brock2018smash,bender2018understanding,liu2018darts,cai2018proxylessnas} draw attention to weight sharing approaches, i.e., One-Shot NAS. These methods first construct an over-parameterized network(also named supernet) that comprises all candidate models, in which architecture weights are trained and shared among different components or models.  The supernet is trained once. SMASH~\cite{brock2018smash} and ENAS~\cite{pham2018efficient} require a hypernetwork or an RNN controller to help generate models. Bender et al.~\cite{bender2018understanding} train the supernet that comprises all candidate paths, and sample models by randomly zeroing out paths. DARTS~\cite{liu2018darts} introduces a real-valued architecture parameter for each path to convert the discrete search space into a differentiable one, then jointly train both architecture parameters and weight parameters. The aforementioned works significantly speed up neural architecture search, however, they suffer from  the large GPU memory consumption issue and hence only afford small-scale tasks, like CIFAR. To save the GPU memory, ProxylessNAS~\cite{cai2018proxylessnas} binarize the architecture parameters and sample only two paths to train in each update step.

More recently, Guo et al.~\cite{guo2019single} propose a single path One-Shot approach, which can be regarded as the extreme way of One-Shot. It gets rid of architecture parameters, and only a single path randomly sampled from the supernet is updated in each step of optimization. After training, an evolutionary algorithm is used to select optimal model. Consequently, it is more flexible and affordable to train and search directly on large datasets(eg., ImageNet). FairNAS~\cite{chu2019fairnas} follows the single path pattern and complies a strict fairness for both sampling and training to alleviate the rich-get-richer phenomenon. ScarletNAS~\cite{chu2019scarletnas} introduces Linearly Equivalent Transformation(LET) to enable variable depths, while retaining training stability and accuracy. AtomNAS~\cite{mei2019atomnas} takes advantage of MixNet~\cite{tan2019mixconv} and further expands to searchable channel proportion of MixConv in an end-to-end form.
PC-NAS~\cite{li2019improving} proposes the posterior fading issue of the weight sharing methods, and introduces a shrinking strategy which guides the proxy distribution to converge towards the true parameter posterior. The difference between PC-NAS and our work is that PC-NAS shrinks the search space layer by layer and finally converges to only several choices of model, while we gradually reduce the remaining number of operations of each layer in a simultaneous manner which makes the contracted supernet still retain a relatively large amount of models.

In conclusion, the above single-path methods are constantly evolving and expanding search space dimensions like kernel size, type of conv, expansion rate, network depth, \textbf{except for} the immutable block channel numbers. Our method releases the block channel number to be searchable and proposes a novel step-shrinking training strategy
 to benefit nerual architecture search.

 \section{Methodology}
In this section, we introduce the proposed approach BS-NAS in a detailed manner.
We first discuss the necessity of channel search as well as its potential obstacles, and define the search space of our supernet from three dimensions: kernel size, expansion ratio and channel number, in Section.~\ref{sec:cs}. Then, based on the supernet constructed in the previous section, we propose a step-shrinking training strategy to deal with the model averaging effects brought by the broadened search space in Section.~\ref{sec:top}. Finally, an evolutionary algorithm is adopted to search the optimal architecture on the shrunk supernet in Section.~\ref{sec:es}.

\subsection{Searchable Channel Number}
\label{sec:cs}
\subsubsection{Necessity and Obstacles}
The number of channels in neural network models plays an important role on the performance of the model. Too many channels could cause model redundancy which reduces model efficiency, and even leads into model overfitting. Contrarily, too few channels are not enough to extract sufficient amounts of discriminative features to fully exploit the potential of the model. Since each layer of the neural network needs to config its number of filters, the problem has exponential complexity as the network goes deeper. These issues become especially prominent on mobile devices which have limited computation resources and tight power budgets. Conventional model construction techniques require domain experts to explore the large design space trading off among model size, speed, and accuracy, which is usually suboptimal and time-consuming.

Some recent works deal with the problem by channel pruning~\cite{he2017channel,he2018amc} or model adaption~\cite{yang2018netadapt,dai2019chamnet}. Such methods usually start with a well-trained neural network model and depend on the weight correlation between different channels on the same layer. It remains a challenging task to search the number of channels in convolutional layers during network designing and training, as the number of output channels in the current layer is correlated with the number of input channels in the next layer. Guo et al.~\cite{guo2019single} propose to utilize a special convolutional kernels, the weights of which (max\_c\_out, max\_c\_in, ksize) are preallocated. Different numbers of weights are sliced out according to the actual needs in current training step, and the optimal number of channels is determined in the evolutionary step. However, they get the best result by running channel search on the best searched model, which can be seemed as a two-stage way. In the proposed approach, we absorb the process of channel search into the training of supernet, and address it in an end-to-end style.

\subsubsection{Revisiting Inverted Residual Block}
Our search space is based on the MobileNetV2's inverted residual block~\cite{sandler2018mobilenetv2}. The inverted residual block contains two pointwise convolutional layers and one depthwise convolutional layer. The first pointwise convolution enlarges the channel number with a expansion ratio to boost the feature diversity, and then the depthwise convolution, which assumes the most important role, calculates the former features to get deeper features in a lightweight way. The last pointwise convolution gathers the features and shrinks to a suitable channel number, without non-linear activation.
\begin{figure*}[t]
\centering
\includegraphics[width=4.8in]{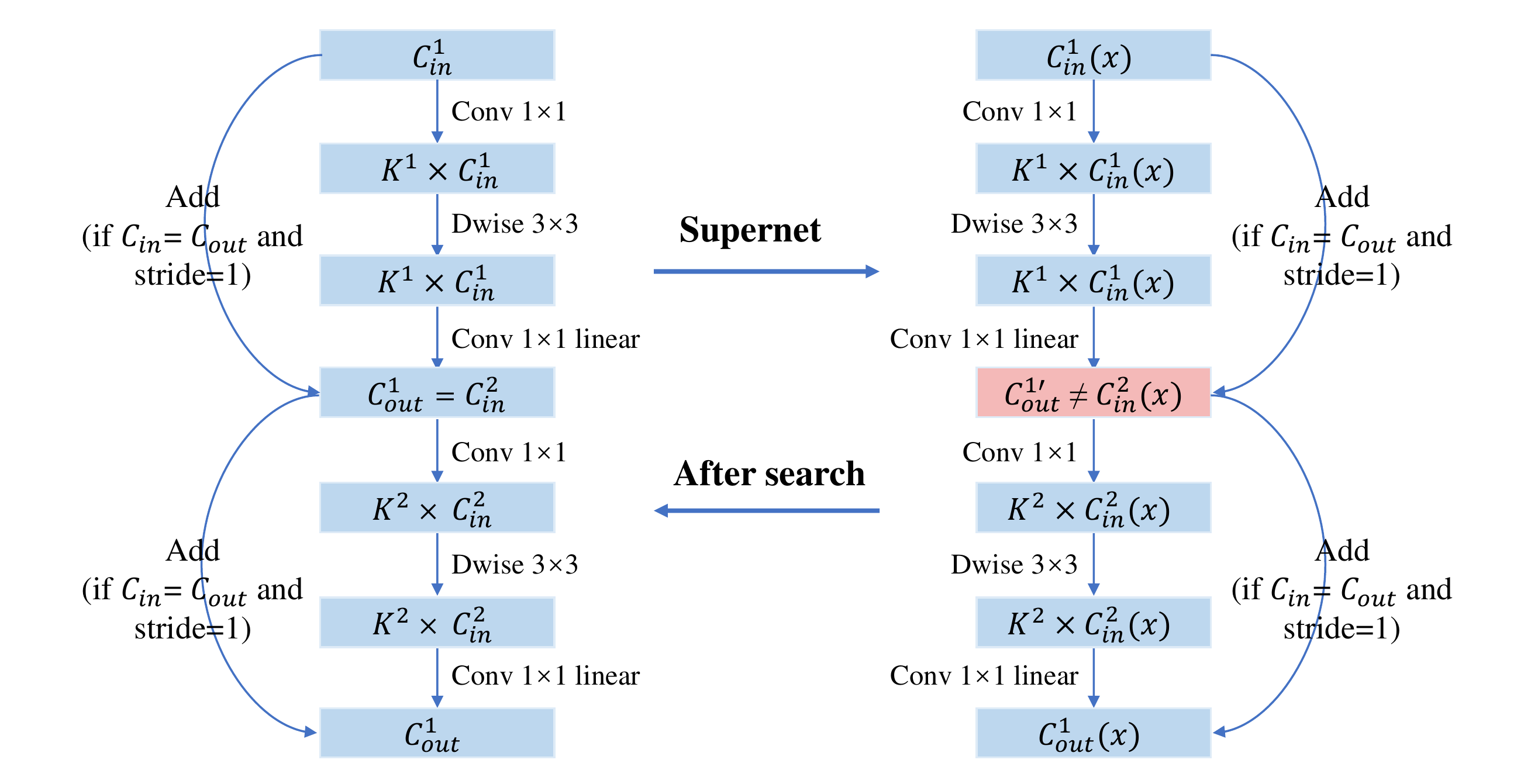}
\caption{The left half describes the two residual blocks stacked together in a MobileNetV2~\cite{sandler2018mobilenetv2} manner, while the right half describes the inconsistency of the input and output between the two residual blocks when the search space contains the number of channels.}
\vspace{-1em}
\label{fig:irb} 
\end{figure*}

Fig.~\ref{fig:irb} shows a simplified network stacked with two inverted residual blocks, $C$ represents the channel number of input or output, $K$ represents expansion ratio. The left part depicts a conventional MobileNetV2 combination, where the output channel number of the first block $C^{1}_{out}$ should be equal to the input channel number of the second block $C^{2}_{in}$. When the number of channels becomes variable which means it has different choices $x$, there comes a contradiction that $C^{1}_{out}(x)$ CANNOT always be equal to $C^{2}_{in}(x)$, as shown on the right part of the figure. The principle is that when the output channel number of the previous block is uncertain, it is impossible to build the next block. Thus, it is inevitable that the number of the output channels $C^{1}_{out}(x)$ must be a determined value. So the key lies in two pointwise convolutional layers, namely the second pointwise convolution of the first block and the first pointwise convolution of the second block. The former transforms an arbitrary input channel number $K^{1} \times C^{1}_{in}(x)$ of the first block to a determined output channel number $C^{1'}_{out}$, and the latter transforms this $C^{1'}_{out}$ to an arbitrary input channel number $K^{2} \times C^{2}_{in}(x)$ of the second block. That is, $C^{1}_{in}(x)$ and $C^{2}_{in}(x)$ can be searched, and the feasibility of the supernet including different channel selections is also guaranteed.

When the supernet training and search is over and the optimal subnetwork is obtained, the number of fixed output channels $C^{1'}_{out}$ is adjusted to the selected number of input channels $C^{2}_{in}(x)$ of the next layer, so that it returns to the normal structure of MobileNetV2 like the left part of Fig.~\ref{fig:irb}. These adjustments happen in the two pointwise convolutional layers. As stated in~\cite{sandler2018mobilenetv2}, the expansion layer acts merely as an implementation detail that accompanies a non-linear transformation of the tensor, while the bottlenecks actually contain all necessary information. Another work~\cite{chu2019scarletnas} proves that the linear pointwise convolution can be used for linearly equivalent transformation and maintain the same representation power for stand-alone models when it is removed.
Nevertheless, it is inevitable to doubt whether the optimal network evaluated from the supernet still has the ability to reliably reflect the performance of the adjusted network. We suppose it is capable, but obviously the less such adjustments the better. The corresponding experiments are provided in Section.~\ref{sec:abs}.

\subsubsection{Constructing Channel Number Searchable Supernet}
\label{sec:ccss}
\begin{table}[]
\vspace{-1em}
\begin{center}
\begin{tabular}{lcccccc}
\hline
No. & Input&Operator&$k$&$t$&$c$&$s$\\
\hline
\multirow{2}*{stem}  &$224^{2}\times 3$&conv2d&3&-&32&2 \\
&$112^{2}\times 32$&separable&3&-&16&1 \\
1-2&$112^{2}\times 16$&bottleneck&(3, 5, 7)&(3, 6)&(24, 32, 40)&2 \\
3-6&$56^{2}\times 40$&bottleneck&(3, 5, 7)&(3, 6)&(40, 48, 56)&2 \\
7-10&$28^{2}\times 56$&bottleneck&(3, 5, 7)&(3, 6)&(64, 72, 80)&2 \\
11-14&$14^{2}\times 80$&bottleneck&(3, 5, 7)&(3, 6)&(96, 112, 128)&1 \\
15-18&$14^{2}\times 128$&bottleneck&(3, 5, 7)&(3, 6)&(160, 192, 224)&2 \\
19&$7^{2}\times 224$&bottleneck&(3, 5, 7)&(3, 6)&(240, 280, 320)&1 \\
\multirow{3}*{tail}&$7^{2}\times 320$&conv2d&1&-&1280&1 \\
&$7^{2}\times 1280$&avgpool $7\times7$&-&-&-&- \\
&$1\times 1\times 1280$&conv2d&1&-&n\_class&- \\
\hline
\end{tabular}
\end{center}
\caption{Channel Searchable Supernet: Each line discribes one layer or cluster with several inverted residual blocks. The Input column is the input size of current layer or the first block of the cluster. The forth and fifth row represent the kernel sizes $k$ and expansion ratios $t$, while the numbers in parentheses represent different choices. The first block of each cluster has a stride $s$ and all others use stride 1. The output channel $c$ is applied equally in the same cluster except the last spring block as discussed.}
\vspace{-2em}
\label{tab:net}
\end{table}
The construction of our supernet is based on the structure of MobileNetV2 as used in~\cite{cai2018proxylessnas,chu2019fairnas}. The number of inverted residual blocks in the network remains unchanged at 19. According to observations, the changes in the number of channels only occur on the 1st, 3rd, 7th, 11th, 15th, and 19th blocks, and the remaining layers are consistent with the number of channels in the previous block, forming six clusters. We retain this way of stacking and only search for the number of channels on these changing blocks, and the remaining blocks maintain a consistent relationship within the same cluster on the number of channels. At the same time, in order to deal with the inconsistency of the input and output caused by the channel search, the output channel number of the last block of each cluster is redirected to a fixed value, as discussed above.
We call this block \emph{spring block}, which means that it can flexibly match different numbers of channels.
That fixed value is eventually set to the maximum number of channels that each cluster can choose to ensure that all choices will not cause information loss.

Following the previous methods~\cite{cai2018proxylessnas,chu2019fairnas,chu2019scarletnas},  various kernel sizes (3, 5, 7) and expansion ratios (3, 6) are also searched in every block. Besides, the choice of newly added channel numbers varies between different clusters. The architecture of the entire supernet is described in detail in Table.~\ref{tab:net}.
We MUST emphasize here that although all clusters are given three channel size options in our work just for simplifying search space and relieving the training overload of the supernet, it is obsolutely free and feasible to offer more options and there is no need to maintain the same number of options between clusters.
After the optimal network is searched from the well-trained supernet, the taut spring blocks are released to turn into a normal one. The stand-alone model will then be trained from scratch to achieve the final performance.

\subsection{Step-shrinking Training}
\label{sec:top}
One-Shot Nas method is always plagued by the problem of weight coupling, which causes model averaging effects and bias the performance evaluation of stand-alone models, as identified in previous works~\cite{bender2018understanding,adam2019understanding,guo2019single,chu2019fairnas}. To mitigate these issues, Guo et al.~\cite{guo2019single} propose single path training with uniform path sampling and Chu et al. apply strict fairness on path sampling. However, these methods do not change the fact that, depending on the size of the search space,  each operation (block unit) in the supernet is shared with more than trillions of paths. Besides, channel size is capable of being searched now, the exponential growth of the search space makes the situation more serious, and also challenges the convergence of the supernet. To address the problem, we propose a progressive strategy to shrink search space which is named Turn-off OPerations (TOPs).

\begin{figure*}[]
\centering
\includegraphics[width=4.8in]{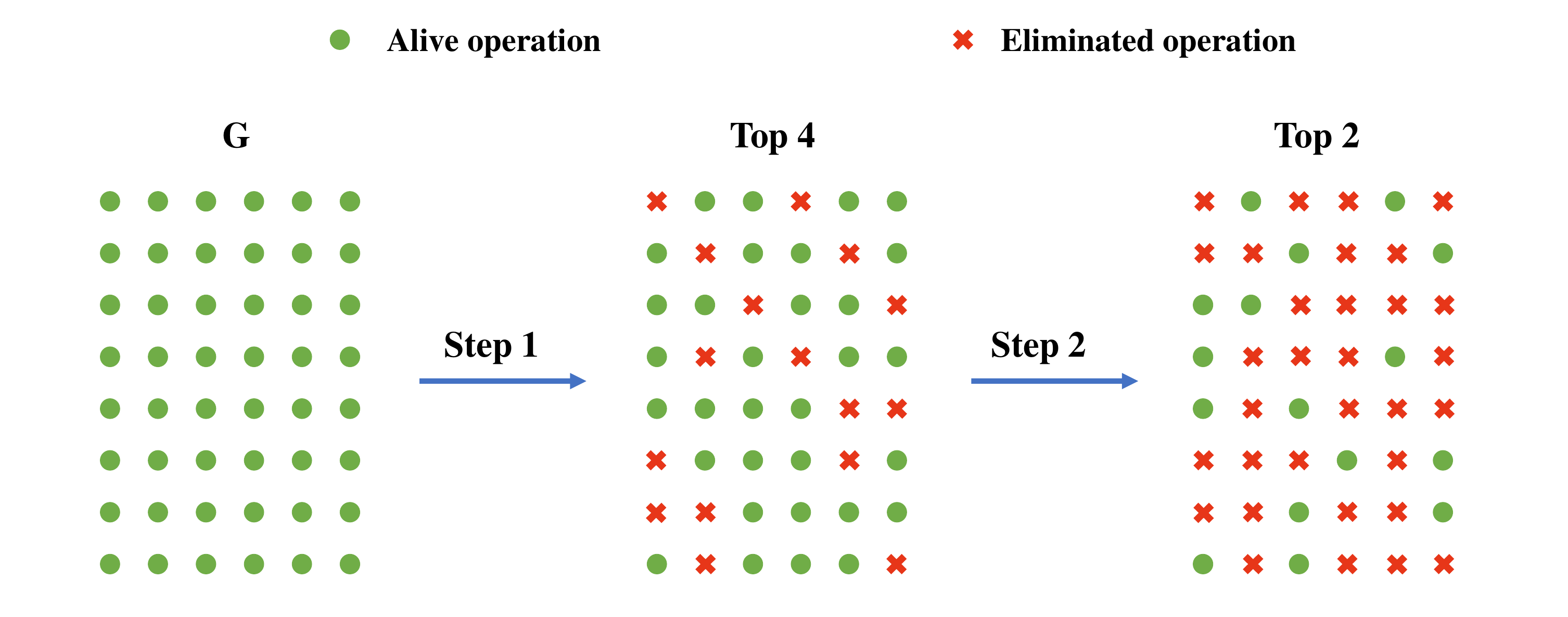}
\caption{Example of the shrinking process of an operational graph. Bright green pot represents an operation which is active for training, red fork represents an operation which is eliminated and will not be chosen in subsequent training. The next contraction changes are based on the previous results.}
\label{fig:shr} 
\end{figure*}

We divide the training of the supernet into two stages. In the first stage, single-path training is performed on all alternative operations. It takes $T_{f}$ epochs to achieve the second stage. The second stage consists of three shrinking steps, each step separated by $T_{i}$ epochs. Figure.~\ref{fig:shr} shows the simplified version of shrinking process. Now imagine that the current supernet is a graph $G$, and each optional operation is regarded as a small highlight on the graph.
\begin{algorithm}[t]
\caption{Step-shrinking Training}
\label{algo:1}
\begin{algorithmic}[1] 
\Require supernet $S$, graph of alternative operation $G$, layer number $H$, number of operations $O_{N}$, maximum retention number in each step $N_{r}$, the ith operation $O_{i}$, multiple $M$, accuracy on validation set $Top1\_acc$, importance ratio $R_{i}$, internal between shrinking steps $T_{i}$, first-stage epochs $T_{f}$, model fetched from supernet $m$
\Ensure well-trained supernet in shrunk search space 
\State Random initialization of parameters, $N_{r} = O_{N}$
\State Train $S(G)$ for  $T_{f}$ epochs with fair sampling
\For{$epoch = T_{f} : T_{f}+3T_{i}$}
  \If{$mod(epoch-T_{f}, T_{i}) == 0$}
    \For{$0 : N_{r} \times M$}
      \State Fetch one model $m$ by fair sampling
      \State Evaluate($Top1\_acc, m$)
    \EndFor
    \State Sort($\{m\}, Top1\_acc$)
    \State Reduce $N_{r}$
    \For{$0 : H$}
      \State Calculate($R_{i}$)
      \State Sort($\{O_{i}\}, R_{i}$)
      \State Keep the top $N_{r}$ operations $\{O_{1}, O_{2},..., O_{N_{r}}\}$ with positive $R_{i}$
    \EndFor
    \State Update $G(N_{r})$
  \EndIf{}
  \State Train($S(G, N_{r})$)
\EndFor
\end{algorithmic}
\end{algorithm}
The size of this graph is $H \times O_{N}$, $H$ denotes the number of layers and $O_{N}$ denotes the number of operations. Obviously, all points on the graph are alive before the first shrinking step. At the beginning of the first shrinking step, a number of $N_{r} \times M$ models are fetched from the supernet by fair sampling, ensuring that each operation is taken M times.  Here, $N_{r}$ represents the maximum number of operations keeping bright in each layer, it is currently equal to $W$. These models are tested on the validation set and then sorted by top-1 precision. Based on the sorted results above, operations are sorted in each layer by the following formula:
\begin{eqnarray}
R_{i} = \frac{N_{O_{i}}^{top} - N_{O_{i}}^{bottom}}{N_{O_{i}}},
\end{eqnarray}
where, $N_{O_{i}}$ denotes the total number of occurrences of the operation $O_{i}$, $N_{O_{i}}^{top}$ denotes the number of occurrences in the top third of the models, $N_{O_{i}}^{bottom}$ denotes the number of occurrences in the bottom third of the models, and $R_{i}$ is a ratio which represents the ability of this operation to contribute to high-precision models, the larger the value, the better. In particular, only the operations with positive $R_{i}$ will be considered for retention, unless there is no positive $R_{i}$, the first one will be selected.
Then, we reduce the maximum retention number of $N_{r}$. The top $N_{r}$ operations keep on and the remaining highlights are turned off, which means they will not participate in later training. Subsequently, the supernet trains $T_{i}$ epochs on the updated graph to reach the second step, and repeats the process of the previous step. The whole training process is illustrated in algorithm.~\ref{algo:1}.

Consequently, the search space gradually shrinks as $N_{r}$ gets smaller by steps. Only those operations that consistently perform well can survive the constant elimination. Note that the final number of operations retained in each layer may be less than $N_{r}$ due to the condition that $R_{i}$ must be positive and the channel number choice of blocks in a cluster is limited by its spring block. Thus, the final shrunk search space is much smaller than the initial one, as if the optimal model is locked to a small area, which makes it easier to search for. More importantly, the weight coupling is reduced after the inferior options have been eliminated, and the model averaging effect is alleviated. In particular, both spring block in Section.~\ref{sec:cs} and the shrinking strategy are not limited to the MobileNetV2 framework and can be easily extended to other structures.

\subsection{Evolutionary Search}
\label{sec:es}
 Following~\cite{guo2019single,chu2019fairnas}, an evolutionary algorithm is utilized to search optimal models from the shrunk supernet. This process is not time consuming, as  each model obtained from the supernet inherits its weight and can be tested without finetuning. It is also an efficient way to search in such a huge search space.

In our experiments, we set population size $P = 75$, max iterations $\tau = 20$, and $k = 10$. For crossover, two randomly selected candidates are crossed to produce a new one. For mutation, a randomly selected candidate mutates its every choice block
with probability $\rho = 0.1$ to produce a new candidate. Crossover and mutation are repeated to respectively obtain 25 candidates, the remaining amount is supplemented by random sampling. Unlike the works~\cite{guo2019single,bender2018understanding}, recalculation of the statistics of all Batch Normalization~\cite{ioffe2015batch} operations is not applied before the inference of a candidate model.
 Constraints on model FLOPs (floating point of operations) are also used to generate appropriate candidates throughout the search process.

 \section{Experiments}
\subsection{Experiments Setup}
\subsubsection{Dataset}
We perform all experiments on ImageNet 2012 classification task~\cite{deng2009imagenet} and randomly sample 50,000 images from the train set for validation (50 images per class) during the model search. The original validation set is used as test set to compare the performance with other methods.

\subsubsection{Search Space}
Our search space is constructed based on MobileNetV2~\cite{sandler2018mobilenetv2} as done in~\cite{cai2018proxylessnas,chu2019fairnas}.  We remain the same amount of layers with standard MobileNetV2. Our search dimension contains three directions: kernel size (3, 5, 7), expansion ratio (3, 6) and block channel as discussed in Section.~\ref{sec:ccss}. Squeeze-and-Excitation (SE)~\cite{hu2018squeeze} is not included in our search space. So the size of the search space is $3^{6} \times 6^{19}$, and the FLOPs of these models range from 209M to 812M.

\subsubsection{Training Details}
When training the supernet, we adopt a  stochastic gradient descent optimizer (SGD) with a momentum of 0.9. Linear warmup~\cite{goyal2017accurate} within first five epochs and a cosine learning rate decay strategy~\cite{loshchilov2016sgdr} within a single cycle are applied with an initial learning rate of 0.045. The L2 weight decay is set to $4 \times 10^{-5}$. As discussed in Section.~\ref{sec:top}, the training process contains two stages. The first stage takes $T_{f} = 120$ epochs, and the interval between shrinking steps in the second stage takes $T_{i} = 40$ epochs. Thus, it takes totally 240 epochs to achieve the shrunk supernet. The remaining number $N_{r}$ is orderly set to (18, 9, 5, 3) and the test factor $M$ is 200. Besides, fair sampling~\cite{chu2019fairnas} is also applied to ensure the training fairness.
Extra data augmentations such as MixUp~\cite{zhang2017mixup} and AutoAugment~\cite{cubuk2019autoaugment} are not used because many state-of-the-art algorithms report their results without them.

We use the same hyperparameters as the supernet to train stand-alone models.
Note that unlike some recent works~\cite{chu2019scarletnas,mei2019atomnas} that use a resource-consuming batch size of 4096, all our experiments are done on two Tesla V100 GPUs and the batch size is 512 with the assistance of NVIDIA's mixed precision library Apex. However, large batch size does help improve the performance in some aspects, like statistics of Batch Normalization layers and generalization. Therefore, the performance might not be optimal.

It takes 10 GPU days to training the supernet including shrinking steps. After that, the evolutionary search takes about 10 hours to search for the optimal architecture.

\subsection{Comparison with State-of-the-arts}

Table.~\ref{tab:comp} shows the performance of our model compared with previous state-of-the-arts. In order to make a fair comparison, we only list the results reported without Squeeze-and-Excitation (SE)~\cite{hu2018squeeze} module or extra data augmentations such as MixUp~\cite{zhang2017mixup} and AutoAugment~\cite{cubuk2019autoaugment}. These techniques have been proven effective by numerous experiments that can universally improve the performance of the base model. However, here we focus more on the performance of the network itself.
Our BS-NAS-A achieves 75.9\% top-1 accuracy, 0.8\% and 0.6\% higher than Proxyless (GPU)~\cite{cai2018proxylessnas} and FairNAS-A~\cite{chu2019fairnas} under the same network depth and basic blocks. Especially, BS-NAS-B achieves 76.3\% top-1 accuracy, surpasses all other models, including the latest work AtomNAS-C which takes advantage of Mixconv~\cite{tan2019mixconv} kernels. We must state here that, although our models have slightly larger FLOPs than some methods, one main reason is that large-scale datasets like ImageNet usually encourage wider or deeper models to achieve better performance. And our method just realizes the search of channel numbers, making it easier for the supernet to control the scale of the model to achieve the best performance. In other words, our broadening-and-shrinking search space could adaptively choose the model with appropriate scale based on the size of the dataset.
\begin{table}
\renewcommand\tabcolsep{5.0pt}
\begin{center}
\begin{tabular}{lcccc}
\hline
Model &Parameters &FLOPs &Top-1(\%) &Top-5(\%) \\
\hline
MobileNetV1~\cite{howard2017mobilenets} &4.2M&575M&70.6&89.5\\
MobileNetV2~\cite{sandler2018mobilenetv2} &3.4M&300M&72.0&91.0\\
MobileNetV2 (1.4$\times$) &6.9M &585M &74.7 &92.5\\
ShuffleNetV2~\cite{ma2018shufflenet} &3.5M &299M &72.6 &-\\
ShuffleNetV2 (2$\times$) & 7.4M &591M &74.9 &-\\
\hline
NASNet-A~\cite{zoph2016neural} &5.3M &564M &74.0 &91.6\\
Proxyless (GPU)~\cite{cai2018proxylessnas} &7.2M &465M &75.1 &92.5\\
DARTS~\cite{liu2018darts} &4.9M &595M &73.1 &-\\
FBNet-C~\cite{wu2019fbnet} &5.5M &375M &74.9 &-\\
SPOS~\cite{guo2019single} &-&319M &74.3&-\\
SinglePath~\cite{stamoulis2019single} &4.4M &334M &75.0 &92.2\\
PDARTS~\cite{chen2019progressive} &4.9M &557M &75.6 &92.6\\
FairNAS-A~\cite{chu2019fairnas} &4.6M &388M &75.3 &92.4\\
AtomNAS-C~\cite{mei2019atomnas} &4.7M &360M &75.9 &92.7\\
\textbf{BS-NAS-A (ours)}$\ast$ &4.9M &465M &75.9&92.8\\
\textbf{BS-NAS-B (ours)}$\ast$ &5.2M &613M &76.3&92.9\\
\hline
\end{tabular}
\end{center}
\caption{Comparison with state-of-the-arts on ImageNet under the mobile setting.
$\ast$ denotes searchable number of channels.}
\vspace{-1em}
\label{tab:comp}
\end{table}

\begin{figure*}[]
\centering
\includegraphics[width=4.6in]{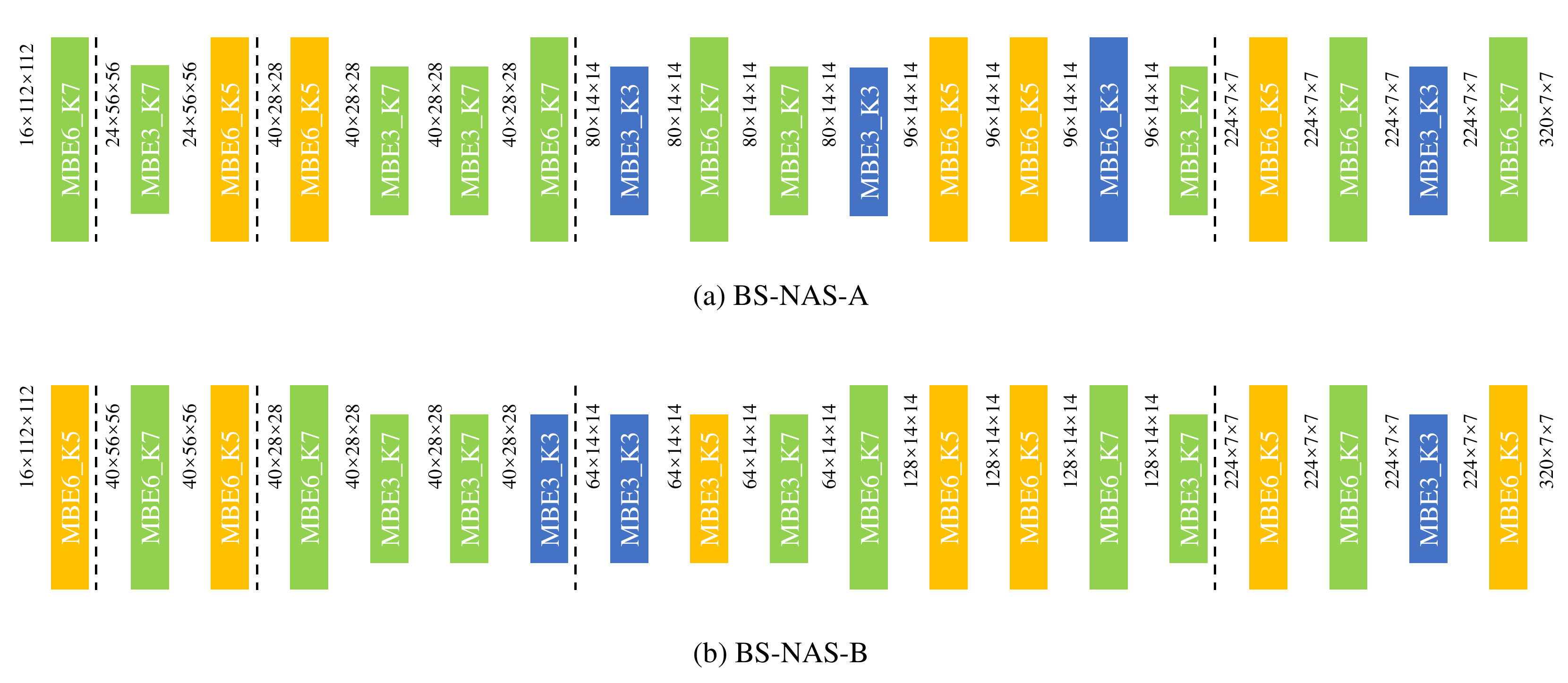}
\caption{The architectures of BS-NAS-A,B. Dashed lines refer to downsampling points. The stem and tail parts
are omitted. }
\vspace{-1em}
\label{fig:mod} 
\end{figure*}

The structures of BS-NAS-A and BS-NAS-B are shown in Figure.~\ref{fig:mod}. Both networks tend to encourage large channel numbers (i.e., select the maximum channel number) in the last several blocks, especially for BS-NAS-B from the 11th block to the end. This is mainly aimed to extract more effective features by increasing the number of channels, thereby overcoming the information loss caused by reduction of resolutions. They also share the same structure from the 15th to 18th blocks, which strongly proves that our shrinking strategy can effectively retain well-performing structures of an interrelated region. Note that BS-NAS-B selects the maximum channel number in the first two blocks and keeps the channel number after the first downsampling, which is a little bit counterintuitive. It seems that increasing the number of channels in the first two layers to get more information is more worthwhile than remedially increasing the number of channels after downsampling.

\subsection{Algorithmic Study}
\label{sec:abs}
\begin{figure*}[]
\centering
\includegraphics[width=4in]{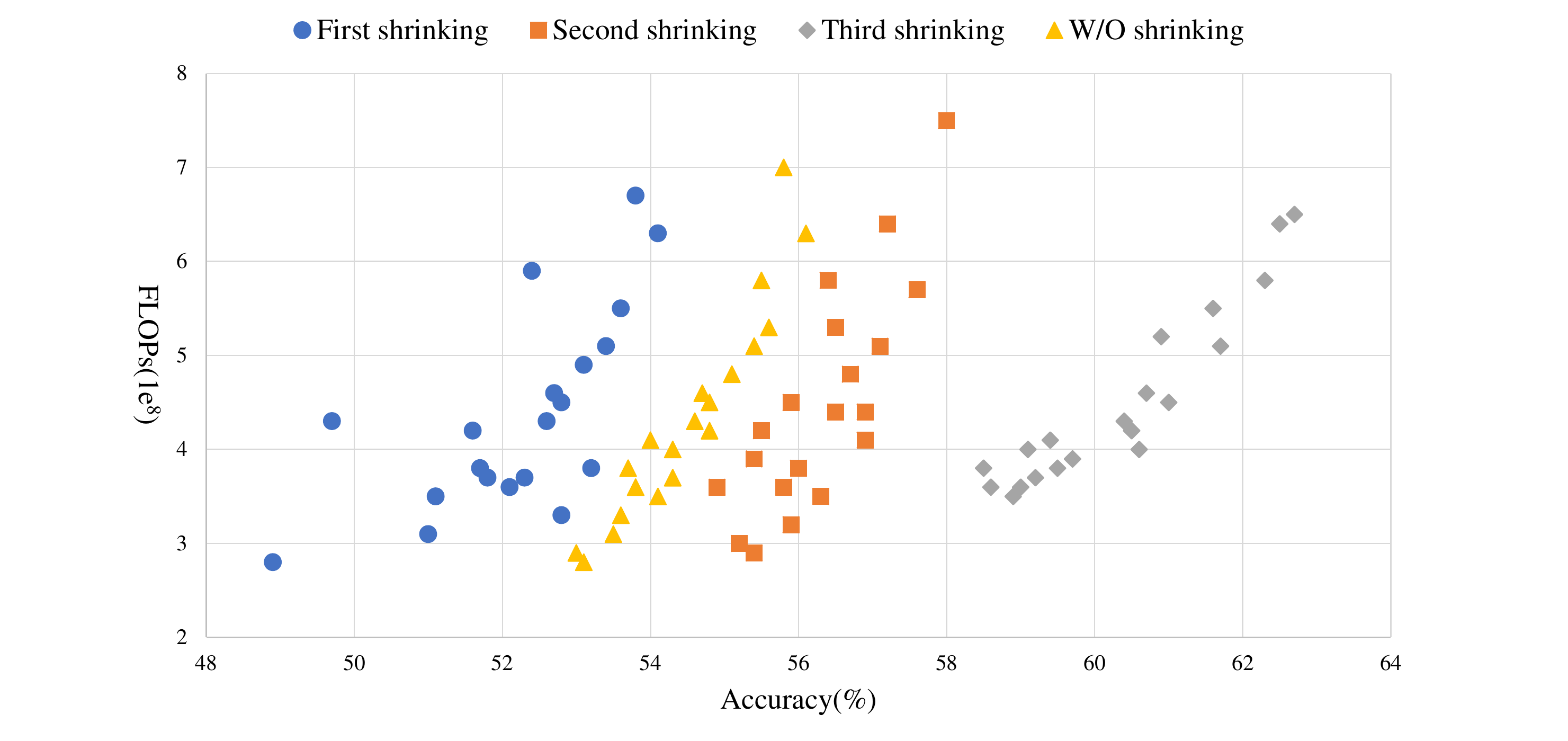}
\caption{Top-1 accuracy distribution of One-Shot models randomly sampled from three shrinking steps and normal training without shrinking respectively. }
\vspace{-1em}
\label{fig:cur} 
\end{figure*}

In this section, we perform a study to demonstrate the effectiveness of the broadening-and-shrinking manner.
We randomly sample 20 One-Shot models from three shrinking steps and normal training 240 epochs without shrinking respectively, and evaluate these models on the ImageNet validation set without recalculation of statistics for Batch Normalization. Top-1 accuracy distributions of these four categories are illustrated in Figure.~\ref{fig:cur}. As the shrinking step progresses, the accuracy distribution improves overall. This is not only because the training epoch is increasing, but also because the search space is continuously shrinking, which means that the remaining well-performing operations have more opportunities to participate in training. In addition, with the same number of trainings, models without shrinking have much lower accuracy than models from the last shrunk search space, even inferior to models after the second shrinkage with 40 epochs less training. It is further observed that the accuracy distribution of models after the third shrinkage has a wider range from 58.2 to 62.3 on the accuracy axis than the accuracy distribution of training without shrinking. This indicates that our shrinking strategy effectively alleviate the model averaging effects, which can better highlight the performance differences between different models and help the final selection of the optimal model.

Besides, we carry out an extreme experiment in which we set the final remaining number $N_{r}$ to 1. Only one operation in each layer is retained, indicating that there is only one model left, no search is required. We train this model from scratch and get a decent top-1 accuracy 75.1\%. Nevertheless, the result is 1.2\% lower than the search result of evolutionary algorithms, mainly because this extreme model retains those operations that are statistically optimal, rather than individual optimal. That's why we keep the shrunk search space within a reasonable scale.

\begin{figure*}[]
\centering
\includegraphics[width=4.2in]{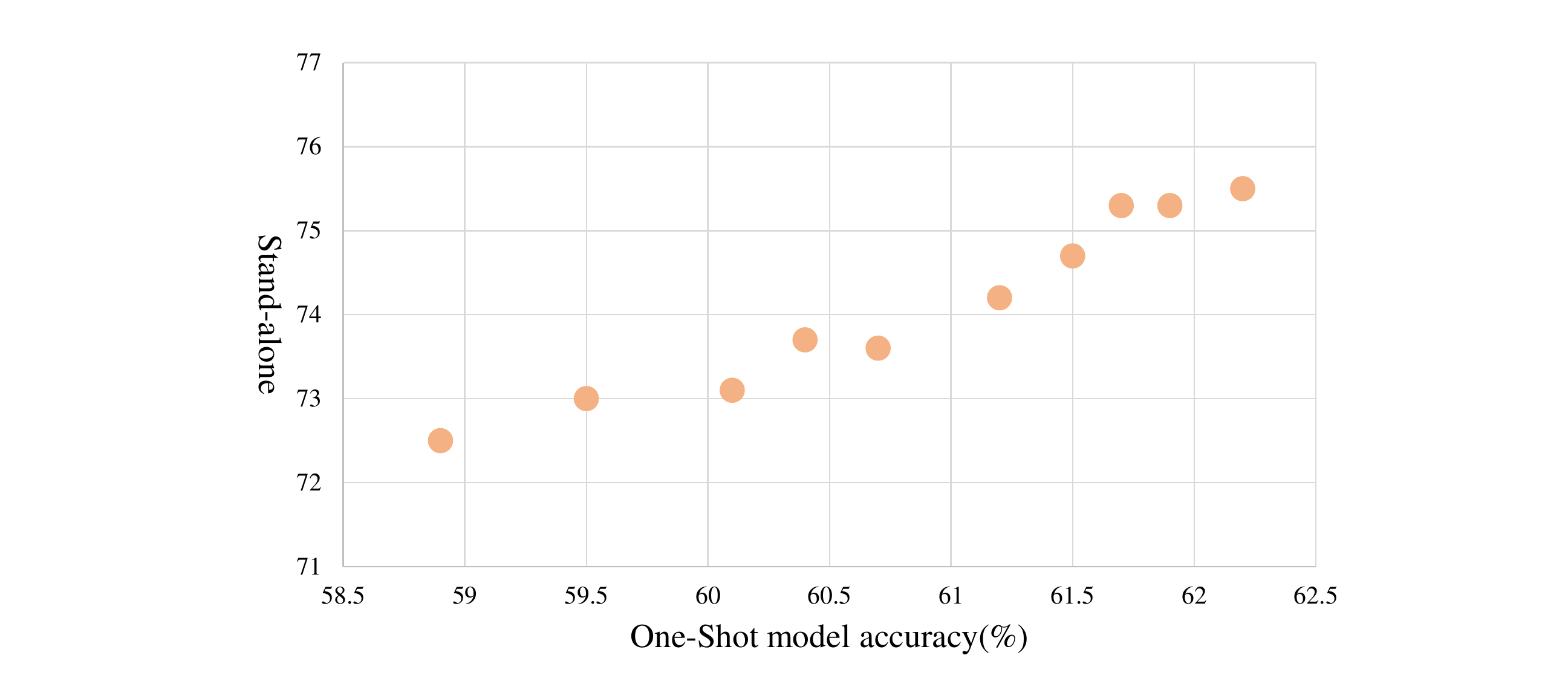}
\caption{Correlation of top-1 accuracies between stand-alone models and One-Shot models.}
\label{fig:pos} 
\end{figure*}

To analyze the correlation between stand-alone models and One-Shot models, we sample 10 One-Shot models with different top-1 accuracies derived from the results of evolution algorithm. Then we train these 10 models from scratch to get the ranking, which is shown in Figure.~\ref{fig:pos}.
We notice that the accuracy gap between stand-alone models and One-Shot models is close to 14\%, probably because the channel number search enriches the search dimension while at the same time increases the difficulty of supernet convergence. Therefore, it is more necessary to accelerate the supernet convergence by shrinking the search space.
In addition, the accuracy of the One-Shot models is nearly positively related to the accuracy of the stand-alone models. A model that performs well in the One-Shot framework still has excellent performance under independent training. It demonstrate that the spring block does not break the correlation between the One-Shot model and the stand-alone model, and the shrunk search space reduces the weight coupling, thereby further improving the ranking correlation.

 \section{Conclusions}
In this paper, we propose a Broadening-and-Shrinking One-Shot NAS (BS-NAS) framework to enrich the representation of the search space meanwhile mitigate model averaging effects. We propose a {\em spring block} to enable One-Shot algorithms search for numbers of channels and a novel {\em shrinking} strategy that gradually eliminates those underperforming operations during training of the supernet. The above end-to-end process is not limited to the MobileNetV2 framework and can be easily extended to other neural structures. Extensive experiments indicate that our BS-NAS achieves the state-of-the-art performance on ImageNet dataset. There exist many effective techniques like Squeeze-and-Excitation, MixUp and AutoAugment in literature. As future work, we will manage to incorporate them into our BS-NAS framework to further improve the performance, and validate on more datasets.

\clearpage
%
%
\bibliographystyle{splncs04}
\bibliography{egbib}
\end{document}